%% file: neurips_2023.tex
\documentclass{article}
\usepackage[preprint]{neurips_2023}

\usepackage[utf8]{inputenc} 
\usepackage[T1]{fontenc}    
\usepackage{hyperref}       
\usepackage{url}            
\usepackage{booktabs}       
\usepackage{amsfonts}       
\usepackage{nicefrac}       
\usepackage{microtype}      
\usepackage{xcolor}         

\newcommand{\thetav}{\boldsymbol{\theta}}
\newcommand{\xval}{x^\text{ho}}

\newcommand{\yval}{y^\text{ho}}

\newcommand{\Dtrain}{\mathcal{D}_\text{t}}

\newcommand{\Dval}{\mathcal{D}_\text{ho}}

\newcommand{\M}{\mathcal{M}(\thetav)}
\newcommand{\Mproxy}{\mathcal{M}_{proxy}(\thetav')}

\usepackage{xcolor}         %
\definecolor{myblue}{rgb}{0.00, 0.45, 0.85}
\definecolor{mygreen}{rgb}{0.05, 0.70, 0.60}
\definecolor{myred}{rgb}{0.84, 0.25, 0.00}
\usepackage{mathtools} 
\usepackage{mathcommand}
\usepackage{amssymb}
\usepackage{nicefrac}

\DeclareMathOperator{\opEntropy}{H}
\DeclareMathOperator{\opPEntropy}{h}
\DeclarePairedDelimiterXPP{\Hof}[1]{\opEntropy}{[}{]}{}{
    
    \ifblank{#1}{\:\cdot\:}{#1}
}
\DeclarePairedDelimiterXPP{\hof}[1]{\opPEntropy}{[}{]}{}{
    
    \ifblank{#1}{\:\cdot\:}{#1}
}
\DeclarePairedDelimiterXPP{\Lof}[1]{L}{[}{]}{}{%
    
    \ifblank{#1}{\:\cdot\:}{#1}
}
\DeclarePairedDelimiterXPP{\htof}[1]{\opPEntropy_{\text{true}}}{[}{]}{}{%
    
    \ifblank{#1}{\:\cdot\:}{#1}
}
\DeclarePairedDelimiterXPP{\xHof}[1]{\opEntropy}{(}{)}{}{
    \ifblank{#1}{\:\cdot\:}{#1}
}

\DeclareMathOperator{\opMI}{I}
\DeclarePairedDelimiterXPP{\MIof}[1]{\opMI}{[}{]}{}{
    
    \ifblank{#1}{\:\cdot\:}{#1}
}
\DeclareMathOperator{\opPMI}{pmi}
\DeclarePairedDelimiterXPP{\PMIof}[1]{\opPMI}{[}{]}{}{
    
    \ifblank{#1}{\:\cdot\:}{#1}
}

\DeclarePairedDelimiterXPP{\CrossEntropy}[2]{\opEntropy}{(}{)}{}{
    \ifblank{#1#2}{\:\cdot\: \MidSymbol[\delimsize\Vert] \:\cdot\:}{#1 \MidSymbol[\delimsize\Vert] #2}
}

\DeclareMathOperator{\opKale}{D_\mathrm{KL}}
\DeclarePairedDelimiterXPP{\Kale}[2]{\opKale}{(}{)}{}{
    \ifblank{#1#2}{\:\cdot\: \MidSymbol[\delimsize\Vert] \:\cdot\:}{#1 \MidSymbol[\delimsize\Vert] #2}
}

\DeclareMathOperator{\opp}{p}
\DeclarePairedDelimiterXPP{\pof}[1]{\opp}{(}{)}{}{
    
    \ifblank{#1}{\:\cdot\:}{#1}
}
\DeclarePairedDelimiterXPP{\ptof}[1]{\opp_{\text{true}}}{(}{)}{}{
    
    \ifblank{#1}{\:\cdot\:}{#1}
}
\DeclarePairedDelimiterXPP{\pevalof}[1]{\opp_{\text{eval}}}{(}{)}{}{
    
    \ifblank{#1}{\:\cdot\:}{#1}
}

\DeclarePairedDelimiterXPP{\ppof}[1]{\opp'}{(}{)}{}{
    
    \ifblank{#1}{\:\cdot\:}{#1}
}

\DeclarePairedDelimiterXPP{\pcof}[2]{\opp_{#1}}{(}{)}{}{
    
    \ifblank{#2}{\:\cdot\:}{#2}
}

\DeclareMathOperator{\opq}{q}
\DeclarePairedDelimiterXPP{\qof}[1]{\opq}{(}{)}{}{%
    
    \ifblank{#1}{\:\cdot\:}{#1}
}

\DeclarePairedDelimiterXPP{\xvarHof}[2]{\opEntropy_{\ifblank{#1}{\:\cdot\:}{#1}}}{(}{)}{}{%
    
    \ifblank{#2}{\:\cdot\:}{#2}
}

\title{Irreducible Curriculum\\ for Language Model Pretraining}

\author{%
  Simin Fan\\
  EPFL\\
  \texttt{simin.fan@epfl.ch}\\
  \And
  Martin Jaggi \\
  EPFL \\
  \texttt{martin.jaggi@epfl.ch} \\
}
\usepackage{hyperref}
\usepackage{comment}
\usepackage{graphicx,wrapfig}
\usepackage{amsmath}
\usepackage{amssymb}
\usepackage{algorithm}
\usepackage{algorithmic}

\begin{document}

\maketitle

\input{sections/0-abstract}
\input{sections/1-intro}
\input{sections/2-background}
\input{sections/3-method}
\input{sections/4-experiment}
\input{sections/5-conclusion}

\bibliography{neurips_2023}

\newpage
\input{sections/x-appendix}
\end{document}

%% file: sections/0-abstract.tex
\begin{abstract}
Automatic data selection and curriculum design for training large language models is challenging, with only a few existing methods showing improvements over standard training.
Furthermore, current schemes focus on domain-level selection, overlooking the more fine-grained contributions of each individual training point. It is difficult to apply traditional datapoint selection methods on large language models: most online batch selection methods perform two-times forward or backward passes, which introduces considerable extra costs with large-scale models. 
To mitigate these obstacles, we propose \textsc{irreducible curriculum} as a curriculum learning algorithm for language model pretraining, which prioritizes samples with higher learnability. Specifically, to avoid prohibitive extra computation overhead, we simulate the sample loss along the main model's training trajectory using a small-scale proxy model. Our experiments on the RedPajama-1B dataset demonstrate a consistent improvement on validation perplexity across all 7 domains compared to random uniform baseline and the anti-curriculum strategy. Our method also reduces the sharpness of the network and illustrates a better 5-shot accuracy on MMLU benchmarks.
\end{abstract}

%% file: sections/1-intro.tex
\section{Introduction}
The evolution of language models demonstrates impressive generation and reasoning abilities from a continuous scaling-up of model size and training corpus~\citep{brown2020language, chowdhery2022palm, kaplan2020scaling,hoffmann2022training}, along with very significant growth in computation cost. Besides the quantity~\citep{kaplan2020scaling} and quality~\citep{lee2022deduplicating,longpre2023pretrainers} of training data, several recent works show that the ordering~\citep{chen2023skillit} and composition~\citep{xie2023doremi} of data during the training can highly impact performance and efficiency. \cite{xie2023doremi} proposes \textsc{DoReMi}, which determines the optimal domain mixture to construct the pretraining corpus using auxiliary models; \cite{chen2023skillit} introduces an online selection scheme, which seeks to dynamically update the mixture of data from each skill-set at each training step. While \textsc{DoReMi} and \textsc{Skill-it} demonstrates great downstream performance, their curriculum are built upon the group-level where all the instances within one group (domain/skill) shares the same sampling probability, without looking inside each domains on the sample-specific attributions. 

Nevertheless, few research have succeeded applying traditional sample-level selection schemes for large langauge model pretraining. 
Online batch selection methods~\citep{loshchilov2015online, katharopoulos2018not, jiang2019accelerating, schaul2015prioritized} select hard samples with high loss or high gradient norm, which require a second forward/backward pass. That introduces large extra computation costs when the model size is large, which hurts the scalability. On the other side, \citet{campos2021curriculum} shows that the linguistic-based curriculum learning failed to improve on causal language model pretraining. 
\citet{schaul2015prioritized} introduces \textsc{Rho-Loss}, which is an online batch selection scheme based on the gap between current training loss and an irreducible loss term measured using a proxy model. Inspired by the idea of proxy model approximation, we propose \textsc{irreducible curriculum} as a curriculum learning algorithm, construct an ordered data stream by simulating the learning trajectory of large model through the lens of a small-scale proxy model. Specifically, we train the small-scale proxy model on a holdout set, and measure the loss on each data samples in the training set using the proxy model at different training stages. We estimate the \emph{learnability score} of the training samples by the loss gap from the proxy model at early- and late-stage. The \textsc{irreducible curriculum} would prioritize the samples with highest \emph{learnability score} and proceed to the samples with lower \emph{learnability score}. We present the details of our methodology in Section \ref{sec:method}.

%% file: sections/2-background.tex
\section{Background}
\subsection{Reducible Holdout Loss Selection (\textsc{Rho-Loss})}
We recapitulate \textsc{Rho-Loss}~\citep{mindermann2022prioritized} with the emphasis on their data selection criterion and approximation using the proxy model. Considering a model $\M$ parameterized by~$\thetav$, and training corpus $\mathcal{D}_{train} = \{(x_i, y_i)\}_{i=1}^n$, online batch selection pre-samples a larger batch $B_t \in \mathcal{D}_{train}$ uniformly, from which it then selects a smaller subset $\displaystyle{b_t \in B_t}, \displaystyle{|B_t| \gg |b_t|}$. 

At training step $t$, given the model has seen part of training corpus $\Dtrain$, the output distribution from $\M$ for an input $x$ is denoted as $p(y|x;\Dtrain)$. Therefore, the log-likelihood objective on the holdout evaluation set after adding a new training point $(x,y)$ can be written as $\log \pof{\yval | \xval; \Dtrain \cup (x, y)}$. Applying Bayesian rule and conditional independence, the notion of the \textsc{Rho-Loss} can be derived as,
\begin{align} 
    &\log \pof{\yval | \xval; \Dtrain \cup (x, y)}\notag\\[0.5ex]
    & = \log \frac{ \pof{y | x; \xval, \yval, \Dtrain} \pof{\yval | \xval, x; \Dtrain} }{ \pof{y | x, \xval; \Dtrain}}  
    \text{\begin{tabular}{c}
        Bayes rule  
    \end{tabular}}\notag
    \\[0.5ex]
    & = \log  \frac{ \pof{y | x; \yval, \xval, \Dtrain} \pof{\yval | \xval; \Dtrain} }{ \pof{y | x; \Dtrain} } 
    \text{\begin{tabular}{c}
        conditional\\
        independence
    \end{tabular}}\notag \\[0.5ex]
    & \propto L\left[y|x;\thetav[t],\Dtrain\right] - L\left[y|x;\thetav[t],\Dval,\Dtrain\right] \notag \\
    & \approx L\left[y|x;\thetav[t],\Dtrain\right] - L\left[y|x;\thetav[t],\Dval\right] \label{equ:rho_loss} \\
    & \approx \overbrace{\underbrace{L\left[y|x;\thetav[t],\Dtrain\right]}_{\text{training loss}}  ~~~-\underbrace{L\left[y|x;\thetav',\Dval\right]}_{\text{irreducible holdout loss (IL)}}}^{\text{reducible holdout loss}} \text{\begin{tabular}{c}
        proxy-model\\
        approximation
    \end{tabular}}\label{equ:rho_loss_approx}
\end{align},
where $\thetav'$ denotes a smaller scale proxy model pretrained on holdout set $\Dval$.
To avoid retraining the model on $(\Dtrain \cup \Dval)$ at each step, the second term in Equ. \eqref{equ:rho_loss} is approximated with the model trained on the holdout set $L\left[y|x;\thetav[t],\Dval,\Dtrain\right] \approx L\left[y|x;\thetav[t],\Dval\right] \textit{ (irreducible holdout loss)}$, such that the second term could be computed by a pretrained, reusable model. To further improve computational efficiency, the \emph{small-scale proxy model} $\Mproxy$ is applied to compute the irreducible holdout loss. \citet{mindermann2022prioritized} empirically proved that the approximation can preserve the desired properties with significantly lower computational costs.

Qualitatively, \textsc{Rho-Loss} down-weighs the \emph{redundant}, \emph{noisy} samples and \emph{outliers}: redundant points have low training loss and low irreducible holdout loss, while noisy samples and outliers have high training loss and high irreducible holdout loss, both lead to a low \textsc{Rho-Loss} score. While the approximation alleviates the computational burden, the first term in Equ.~\eqref{equ:rho_loss_approx} still requires one forward pass through the proxy model, and two forward passes through the large model at every step: On a pre-sampled batch $B_t$, the scoring takes one pass through the proxy model and the large model. Then the large model would be trained with a second pass on the top-ranked $|b_t|$ with highest scores.
Given that it takes $C_1$ FLOPs for one forward pass through the large model $\M$ and $C_2$ for proxy model $\Mproxy$. To train the model for $T$ steps on $T\cdot |b_t|$ samples, \textsc{Rho-Loss} takes $\mathcal{O}(TC_1\cdot (|B_t|+|b_t|)+TC_2|B_t|)$ forward FLOPs. In case that $\displaystyle{|B_t| \gg |b_t|}$, the extra computation overhead would be significant.\label{sec:background}

%% file: sections/3-method.tex
\section{Irreducible Curriculum}\label{sec:method}
Aiming to reduce the computation cost while improving the training efficiency and robustness, we tend to design a pre-defined curriculum to simulate the online assessment of the scores $\Lof{y | x; \Dtrain}$ along the training process. We refer to our algorithm as \textsc{irreducible curriculum}, since it is determined only by a proxy-model trained on a holdout dataset without resorting any large model's training dynamics. 

Specifically, we train a small proxy model $\Mproxy$ on holdout set $\Dval$ for $T$ overall steps while saving an early-stage checkpoint at $t_0$ steps ($\Mproxy[t_0]$) and a late-stage checkpoint($\Mproxy[T]$). We define the \emph{learnability score} for every sample $(x,y) \in \Dtrain$ of the training set as the gap between the loss from early-stage and the late-stage proxy model checkpoints\footnote{we omit writing $y$ here, as for language modelling the labels $y$ refers to next-token prediction on the current sequence $x$.}:
\begin{equation}
    Learnability(x) := L_{early}-L_{late} = L\left[x;\thetav'[t_0]\right] - L\left[x;\thetav'[T]\right]
    \label{equ:learnability}
\end{equation}
To differentiate better among data points, we take the average of loss from last three checkpoints as the late-stage loss. 

The initial threshold of learnability is set as the top-$\lambda_0$\% instance's score among the training set. The first training batch would be uniformly selected from these top-$\lambda_0$\% instances whose learnability scores are above the threshold. To simulate the decreasing training loss in online data selection (Equ. \ref{equ:rho_loss_approx}), the threshold of learnability is decreasing along the training process following a step function. Specifically, consider that we are applying the curriculum on model $\M$ during the first $T_c$ steps (train for $T$ steps in total), at step $t<T_c$, the training batch for $\M$ is uniformly sampled from $(\lambda_0+(1-\lambda_0)t/T_c)$ instances from the training set with highest learnability. Thus, the sampling probability distribution across training set $\Dtrain$ is:
\begin{align}
    &\mathcal{P}(t) = Unif\left(\{learnability(x)>S(t), x\in \Dtrain\}\right),\\
    &\textit{where  } S(t) \textit{   satisfies   } \frac{|\Dtrain>S(t)|}{|\Dtrain|}=\lambda_0+(1-\lambda_0)t/T_c. \notag
\end{align}
After $T_c$ steps, the uniform random sampling would be applied on the whole dataset.
Our proposed \textsc{irreducible curriculum} introduces no extra forward passes through the large language model $\M$ when we pre-compute the learnability score using the proxy model. Instead, the only extra overhead comes from the proxy model as $\mathcal{O}(C_2\cdot (|D_t|))\geq\mathcal{O}(T_c C_2\cdot (|b_t|))$ forward FLOPs, which could be negligible comparing with $\mathcal{O}(TC_1\cdot (|B_t|+|b_t|)+TC_2|B_t|)$ overhead from \textsc{Rho-Loss}\footnote{$C_1, C_2, B_t, b_t$ notations are kept consistent with Sec. \ref{sec:background}}.

%% file: sections/4-experiment.tex
\section{Experiments and Results}
In this section, we present our initial inquiry experimental results on RedPajama-1B dataset, including~7 domains\footnote{Arxiv, Wikipedia, Book, Common-Crawl, C4, Stackexchange, Github.}. We randomly split each domain dataset into two subsets $\mathcal{D}_{proxy}$ and $\mathcal{D}_{train}$. $\mathcal{D}_{proxy}$ is used to train the proxy model, which is applied to score $\mathcal{D}_{train}$ and construct the curriculum. 
Considering a severe unbalance in domain scale, we uniformly sample from each domain, while applying our proposed \textsc{irreducible curriculum} inside each domain. 
We also provide the results applying the curriculum on the whole dataset in Section \ref{sec:discussion}. 

We use a 6-layer (82M) model as the proxy model $\Mproxy$ to construct the curriculum, which is applied to train a 12-layer (124M) model $\M$. Both the proxy model and the full size model follow the decoder-only transformer architecture  \citep{vaswani2023attention} and share the same set of training hyperparameters (batch size 64, context length 512, learning rate 3e-4). We compare \textsc{irreducible curriculum} (\textsc{Curriculum}) with random uniformly sampling baseline (\textsc{RS}) and anti-curriculum (\textsc{Anti-Curriculum}), where we start with a fraction of $\lambda_0$ samples with lowest score, and then increase the sampling upper bound with the same rate. 
\subsection{Result and Analysis}
\paragraph{\textsc{Irreducible Curriculum} consistently improve downstream performance on all domains.}\label{sec:ppl-result}
\begin{figure}[!ht]
	\centering
	\includegraphics[width=\textwidth]{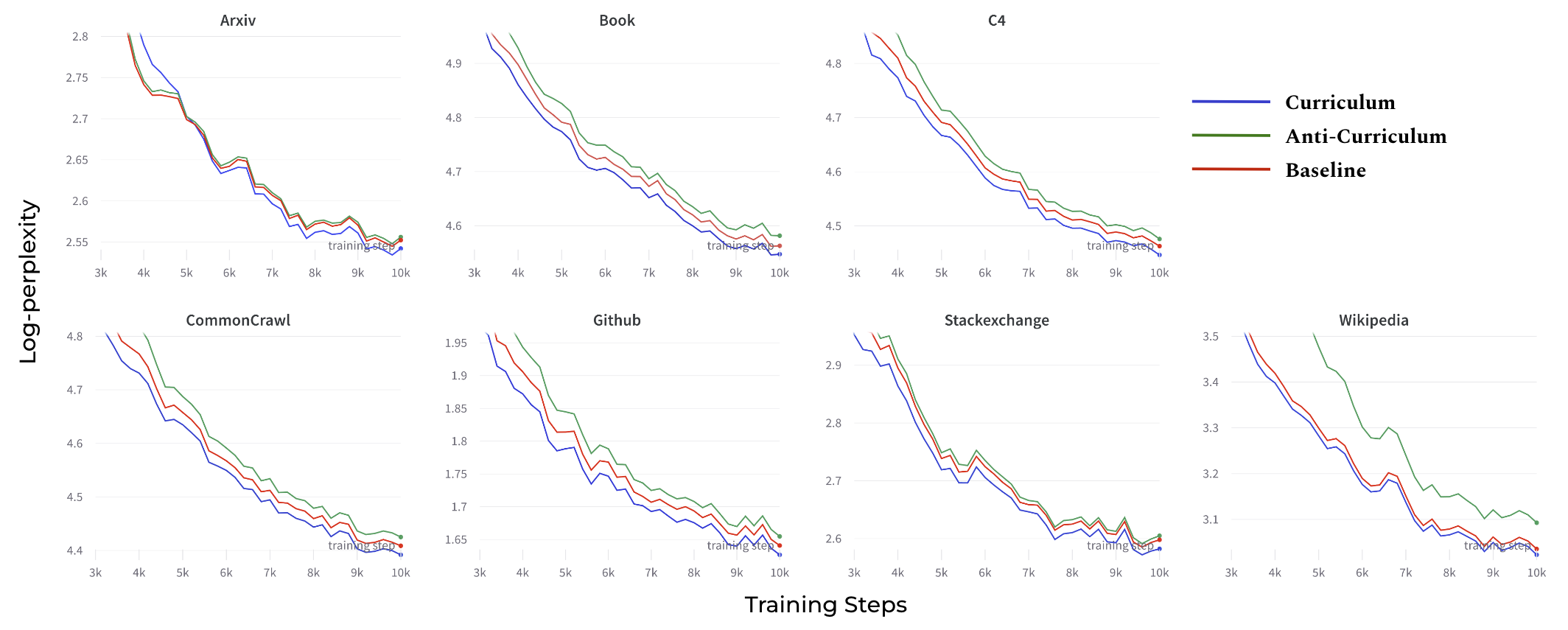}
    \vspace{-15pt}
    \caption{Domain-wise Validation Log-Perplexity on RedPajama-1B. It demonstrates \textcolor{myblue}{\textsc{irreducible curriculum}} consistently outperforms \textcolor{myred}{\textsc{Baseline}} and \textcolor{mygreen}{\textsc{anti-curriculum}}.}
	\label{fig:domain-ppl}
\end{figure}
We apply the irreducible curriculum across every domain within the RedPajama-1B\footnote{https://huggingface.co/datasets/togethercomputer/RedPajama-Data-1T-Sample} dataset, with $\lambda_0 = 50\%$, early stage checkpoint is saved at $t_0=2000$ steps. We train the proxy model (82M) and full model (124M) for $T=10'000$ steps. Since the large model would learn faster than the proxy model, we set $T_c=5000<T$ to compensate the learning pace. 
According to Figure~\ref{fig:domain-ppl}, \textsc{irreducible curriculum} \textbf{\emph{consistently reaches a lower validation perplexity across all 7 domains}} compared with the uniformly random baseline, even trained on a relatively small dataset. Conversely, the \textsc{anti-curriculum} approach leads to a deterioration in the evaluation performance across all domains.

We also perform experiments by varying the values of $\lambda_0$ and $T_c$ (Table \ref{tab:mmlu}). Notably, all the settings applying \textsc{irreducible curriculum} outperform the random uniform sampling baseline, except for the case of ($\lambda_0=25\%, T_c=10'000$). A possible explanation for this is that the step function determined by the proxy model inadequately addresses the differing learning paces between the large and small proxy models. On the contrary, all experiment settings in the \textsc{anti-curriculum} group yield worse validation perplexity results.

Additionally, the empirical results shows that both \textsc{irreducible curriculum} and \textsc{anti-curriculum} approaches outperform the baseline on the 5-shot inference accuracy on MMLU benchmark~\citep{hendrycks2021measuring}. Specifically, ($\lambda_0=25\%, T_c=5000$) works best among all the \textsc{curriculum} settings. Besides, it is worth noting that both \textsc{curriculum} and \textsc{anti-curriculum} improve the reasoning capability compared to baseline, which indicates the subset which contributes the most in perplexity not necessarily contains valuable factual knowledge. How to tackle with this mis-alignment towards our final target could also be a future challenge.
\begin{table*}[t]
    \centering
    \small
\caption{The table presents results for different methods evaluated on average 5-shot accuracy on MMLU benchmark and average validation log-perplexity across 7 domains in RedPajama-1B. The best results are in \textcolor{myred}{\textbf{Red}} and all the results outperform the baseline are in \textbf{Bold}. GPT-3 results refer to \cite{hendrycks2021mmlu}.}
\vspace{5pt}
\label{tab:mmlu}
\begin{tabular}{l|c|c|}
\toprule
 \textsc{Method}& 5-shot MMLU Acc.$\uparrow$ (\%) & Avg. Log Perplexity$\downarrow$\\
\midrule
\textsc{RS Baseline} & 22.9 & 3.810\\
\textsc{GPT-3 (6.7B)} & 24.9 & $\slash$\\
\textsc{GPT-3 (13B)} & 26.0 & $\slash$\\
\textsc{GPT-3 (175B)} & 43.9 & $\slash$\\
\midrule
\textsc{Curriculum ($\lambda_0=25\%, T_c=2000$)} & \textbf{26.0} & \textbf{3.799}\\
\textsc{Curriculum ($\lambda_0=25\%, T_c=5000$)} & \textcolor{myred}{\textbf{26.9}} & \textbf{3.802}\\
\textsc{Curriculum ($\lambda_0=25\%, T_c=10000$)} & \textbf{23.5} & 3.840\\
\textsc{Curriculum ($\lambda_0=50\%, T_c=2000$)} & \textbf{23.2} & \textbf{3.800}\\
\textsc{Curriculum ($\lambda_0=50\%, T_c=5000$)} & \textbf{24.7} & \textcolor{myred}{\textbf{3.794}}\\
\textsc{Curriculum ($\lambda_0=50\%, T_c=10000$)} & \textbf{24.7} & \textbf{3.800}\\
\midrule
\textsc{Anti-Curriculum ($\lambda_0=25\%, T_c=2000$)} & \textcolor{myred}{\textbf{26.9}} & 3.819\\
\textsc{Anti-Curriculum ($\lambda_0=25\%, T_c=5000$)} & 24.7 & 3.838\\
\textsc{Anti-Curriculum ($\lambda_0=25\%, T_c=10000$)} & \textcolor{myred}{\textbf{26.9}} & 3.892\\
\textsc{Anti-Curriculum ($\lambda_0=50\%, T_c=2000$)} & \textcolor{myred}{\textbf{26.9}} & 3.814\\
\textsc{Anti-Curriculum ($\lambda_0=50\%, T_c=5000$)} & \textcolor{myred}{\textbf{26.9}} & 3.827\\
\textsc{Anti-Curriculum ($\lambda_0=50\%, T_c=10000$)} & \textbf{24.9} & 3.860\\
\midrule
\bottomrule
\end{tabular}
\end{table*}
\paragraph{\textsc{Irreducible Curriculum} reduces sharpness during training.}\label{sec:eigen-result}
Several recent research shows that applying sharpness aware minimization could effectively improve the generalization ability of the network~\citep{foret2021sharpnessaware}. Thus, we resort to inspect into how the \textsc{irreducible curriculum} could impact the loss landscape by estimating its sharpness along the training trajectory. 
Following \citep{Wu2018HowSS,li2022analyzing}, we adopt the popular definition of sharpness as the largest eigenvalue of the Hessian matrix.
We compute the Hessian of all the dense layers and record its top-eigenvalue every 200 steps along the training process. More details on sharpness computation are provided in Appendix \ref{apd:sharpness}.

\begin{figure}[!ht]
	\centering
	\includegraphics[width=\textwidth]{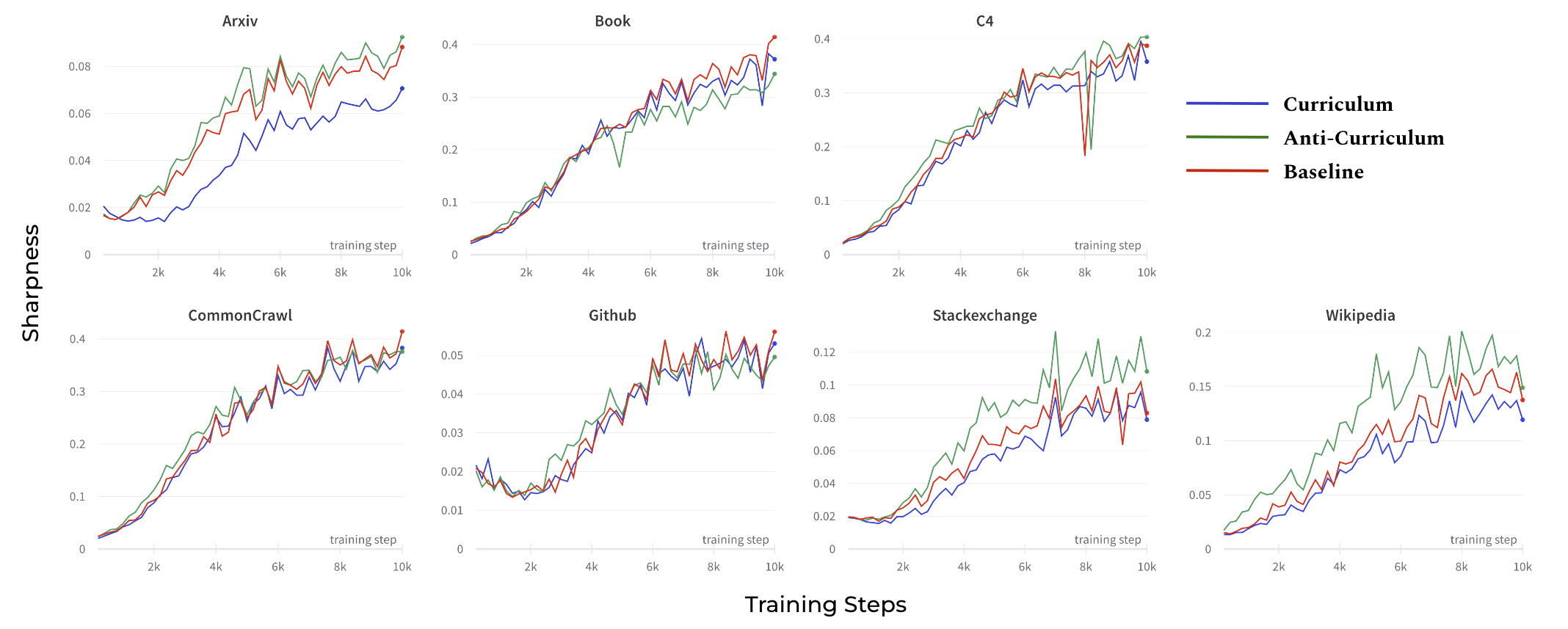}
        \vspace{-15pt}
        \caption{Network sharpness along training process (RedPajama-1B). The sharpness is measured as the largest eigenvalue of the hessian matrix.}
	\label{fig:eigenvalue-redpajama}
\end{figure}

According to Fig. \ref{fig:eigenvalue-redpajama}, training with \textsc{irreducible curriculum} results to a lower sharpness compared with random baseline. On the opposite, the anti-curriculum would lead to a sharper training trajectory. However, whether and how this lower-sharpness correlates with the generalization ability of the network need further explorations on larger scale models and datasets.
\section{Analysis and Failure Mode of Global Curriculum on the Data Mixture}\label{sec:discussion}
Assuming that we have no access to the domain information during training, we can only apply the curriculum on globally on the entire data mixture. 
Instead of starting with top-$\lambda_0$ fraction of samples in each domains, we rank all samples in the global dataset and start sampling from top-scored $\lambda_0$ fraction, which could mostly come from one specific domain. 

\begin{figure}[!ht]
	\centering
	\includegraphics[width=0.95\textwidth]{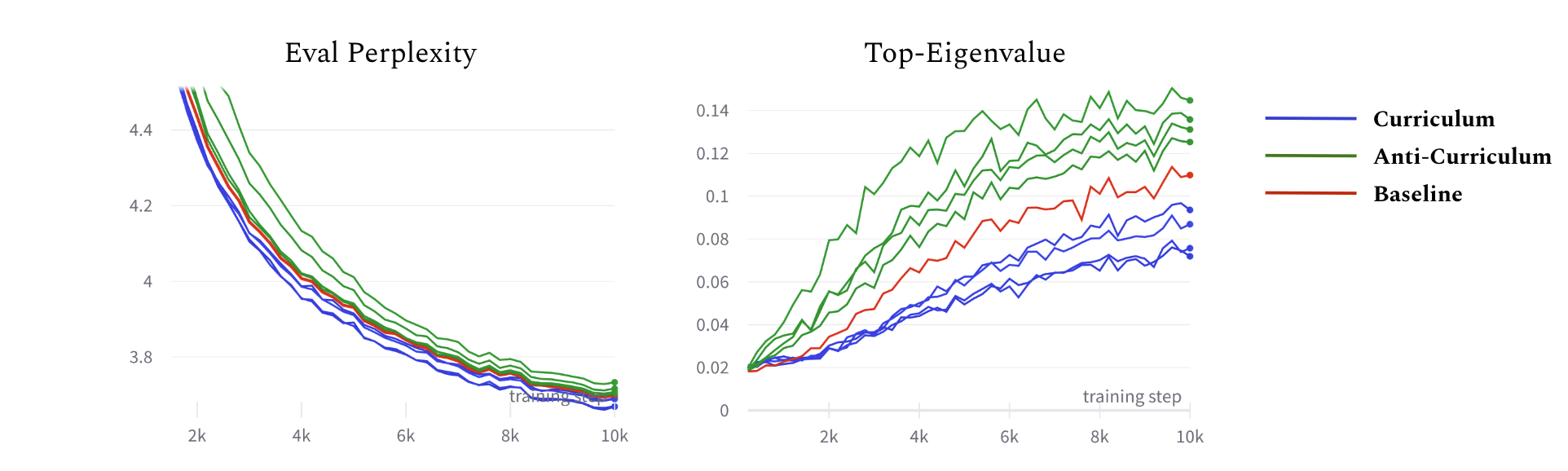}
        \caption{Validation log-perplexity on RedPajama-1B. The validation set is uniformly sampled from all the domains. The \textcolor{myblue}{Blue} lines denotes the \textcolor{myblue}{\textsc{irreducible curriculum}} with different $\lambda_0$ and $T_c$, \textcolor{mygreen}{Green} lines denotes the \textcolor{mygreen}{\textsc{anti-curriculum}}.}
	\label{fig:avg-ppl-mix}
\end{figure}

Here, we compare to the anti-curriculum applied on the whole data mixture, and the baseline with uniformly sampling across all instances. The model scale, architecture, and hyperparameters are the same as in Section \ref{sec:ppl-result}. 
\paragraph{Global curriculum lowers average validation perplexity and reduces sharpness.}
According to Fig. \ref{fig:avg-ppl-mix}, the blue lines denote \textsc{irreducible curriculum} with various combination of $\lambda_0$ and $T_c$\footnote{The combinations are $(\lambda_0, T_c)$: $(25\%, 2000)$, $(25\%, 5000)$, $(50\%, 2000)$, $(50\%, 5000)$.}, while the green lines apply \textsc{anti-curriculum}. Across the entire domain-mixture corpus, the proposed \textsc{irreducible curriculum} could still effectively reduce the validation perplexity and sharpness compared with baseline and \textsc{anti-curriculum} by a large margin. 

\begin{figure}[!ht]
	\centering
	\includegraphics[width=0.9\textwidth]{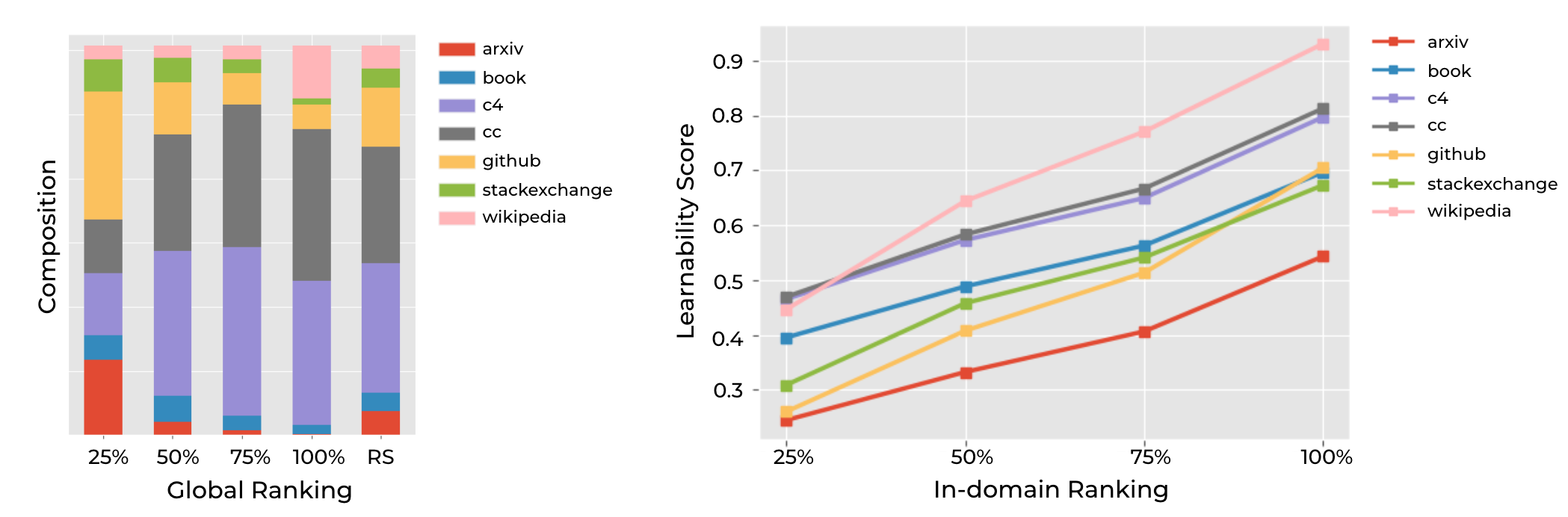}
        \caption{Learnability Scores Distribution. (Left) shows the portion of each domain in each quarter \emph{(0-\textbf{25}\%, 25\%-\textbf{50}\%, 50\%-\textbf{75}\%, 75\%-\textbf{100}\%)} of global ranking; (Right) shows the average learnability score of each quarter in each domain. It demonstrate that the learnability score are not necessarily comparable between different domains because of various lexical distributions and semantic knowledge.}
	\label{fig:domain-composition}
\end{figure}

\paragraph{Global curriculum is not beneficial for every domain.}
We further evaluate the domain-wise performance with the global curriculum applied on the entire corpus. According to Fig. \ref{fig:domain-ppl-mix}, \textsc{irreducible curriculum} reduce the validation perplexity on 4 domains (C4, CommonCrawl, Book and Wikipedia), while showing the opposite effect on the other 3 domains (Arxiv, Github and Stackexchange). 
This can be explained by the domain-wise differences in the learnability score. Figure \ref{fig:domain-composition} shows that most of the globally top-ranked samples are from Wikipedia, C4 and CommonCrawl domains, which would be prioritized by the \textsc{irreducible curriculum} while instances from Github and Stackexchange are mostly within the quarter with lowest scores, which are primarily learnt by the \textsc{anti-curriculum} algorithm. In addition, the in-domain score distributions are highly varying among domains. It illustrates that the an intra-domain learnability (Equ. \ref{equ:learnability}) ranking would offer a more fair comparison, while the naive global curriculum could risk from introducing too much inter-domain sampling bias.
\vspace{-5mm}

%% file: sections/5-conclusion.tex
\section{Future work}
\paragraph{Scaling up the model and training corpus.}
Given the emergency ability of large language models, future work is warranted to explore how this \textsc{irreducible curriculum} can improve the performance of larger scale models, in particular with respect to scaling laws relating dataset size and model size.
\paragraph{Relation between sharpness, generalization ability and curriculum learning.}
In Section \ref{sec:eigen-result}, we observed that the \textsc{irreducible curriculum} could lead to a lower sharpness of the network along the training process while the \textsc{anti-curriculum} approach would lead to the opposite. According to the recent debate on the correlation between sharpness and generalization ability of foundation models~\citep{dinh2017sharp, foret2021sharpnessaware, andriushchenko2023modern}, it is fascinating to uncover whether the curriculum learning algorithm could realize similar outcome as computation-heavy sharpness-aware minimization optimization.

\section{Conclusion}
In this paper, we propose \textsc{irreducible curriculum} as a curriculum learning algorithm with the underlying hypothesis that small proxy model is capable to simulate the early stage training trajectory. For language model pretraining, the method shows a consistent improvement in validation perplexity across several domains and downstream few-shot inference accuracy on the MMLU benchmark. Besides, we also observe our \textsc{irreducible curriculum} can lead to a lower sharpness compared to the random baseline. However, because of the limitation of computation resources, we leave the scaling-up experiments and exploration of further generalization patterns to future investigations. 

%% file: sections/x-appendix.tex
\appendix
\newpage
\section{Preliminary of Hessian and Sharpness}\label{apd:sharpness}
\paragraph{Hessian.}
Considering model parameter $\theta$ and input $x_i$, the hessian is defined as the average second-order derivative of the objective function $\ell$:
\begin{equation}
    \nabla_{\theta}^2\mathcal L(f(\theta))=  \frac{1}{n}\sum_{i=1}^n\nabla_{\theta} ^2\ell(f(\theta, \mathbf{x}_i))
\end{equation}

Specifically, we randomly sample 50 instances from validation set of each domain, and take the average second-order derivative as the hessian matrix. We only consider parameters from all dense layers as $\theta$ while skip the attention matrix.

\paragraph{Sharpness.}
Denote the hessian matrix as $\mathcal{H}$, we adopt a popular definition of the sharpness as the largest eigenvalue of the Hessian $\lambda_{\max}(\mathcal{H})$, following \citep{Wu2018HowSS,li2022analyzing}.

When calculating the sharpness, we apply the numerical method, power method, following \citep{hessian-eigenthings} to calculate the top eigenvalue of the Hessian matrix.

\section{Related Work}
\paragraph{Curriculum Learning for Language Modeling.}
As a popular concept in education and cognitive science, there have been plenty of attempts to apply curriculum learning on training language models, by progressively expose the neural learner with samples with ascending difficulty level \citep{campos2021curriculum,nagatsuka-etal-2021-pre,li2022stabilityefficiency}. However, most of the effective methods increase the difficulty by varying the context length of sequences, without focusing on the syntactic nor semantic properties. \cite{li2022stabilityefficiency} empirically proves that with a incrementally increase context length, the pretraining of language model could be more stable, reducing the number of loss spikes. Similarly, \cite{nagatsuka-etal-2021-pre} shows that feed training samples with increasing block size could improve the performance when pretraining BERT \citep{devlin2019bert}, an encoder-only model with masked language modeling objective.

\cite{campos2021curriculum} has conducted extensive experiments towards using linguistic-featured curriculum learning for language model pretraining. However, the conclusion shows their curriculum fails to effectively help with validation perplexity nor downstream performance.

Recently, \cite{xie2023doremi} has introduced a domain reweighting framework by sampling from domains with various probability. They propose to use two small-scale auxiliary models to determine the domain weights through a group distributional robust optimization game. However, all the samples within one domain would share the same sampling weights, which fails to distinguish the instance-wise attributes inside the domain barrier. How to combine the domain-level reweighting and the in-domain curriculum could also be a promising direction for future exploration.

\begin{figure}[!ht]
	\centering
	\includegraphics[width=0.95\textwidth]{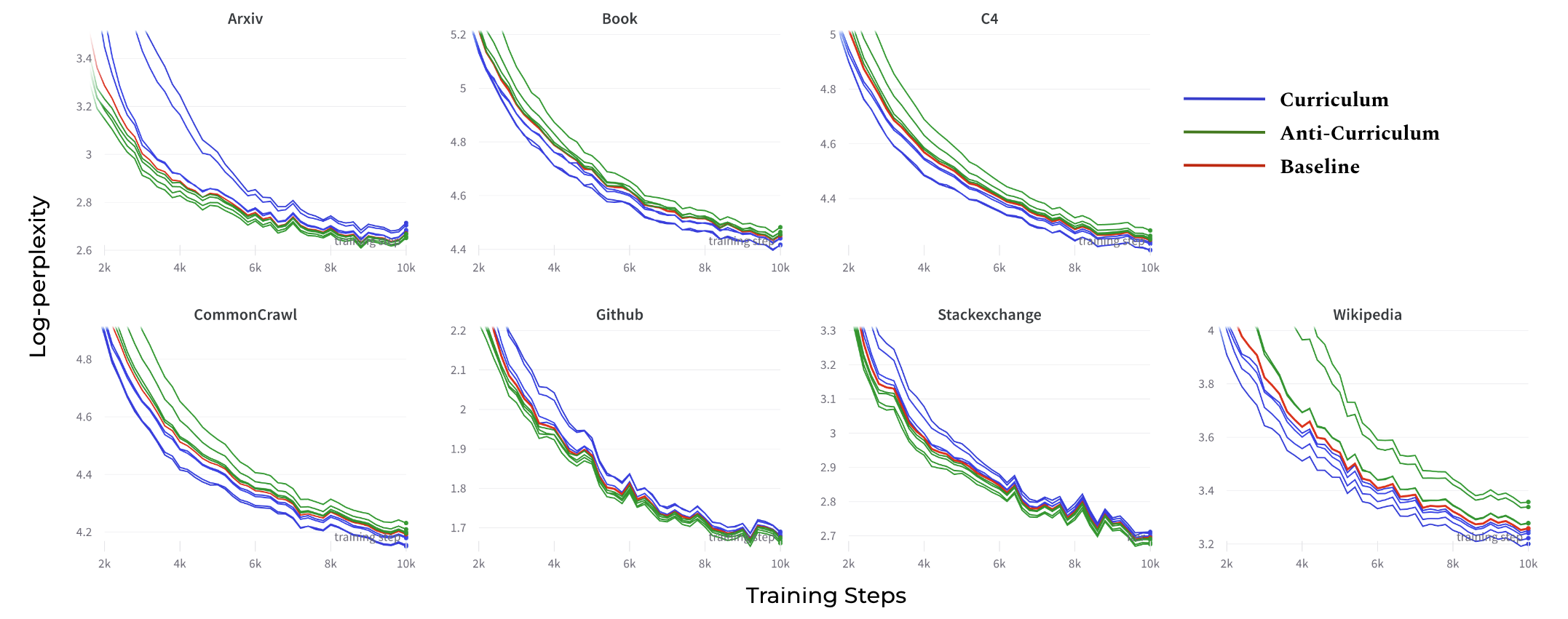}
        \caption{Domain-wise validation log-perplexity on RedPajama-1B.}
	\label{fig:domain-ppl-mix}
\end{figure}

\begin{figure}[!ht]
	\centering
	\includegraphics[width=0.95\textwidth]{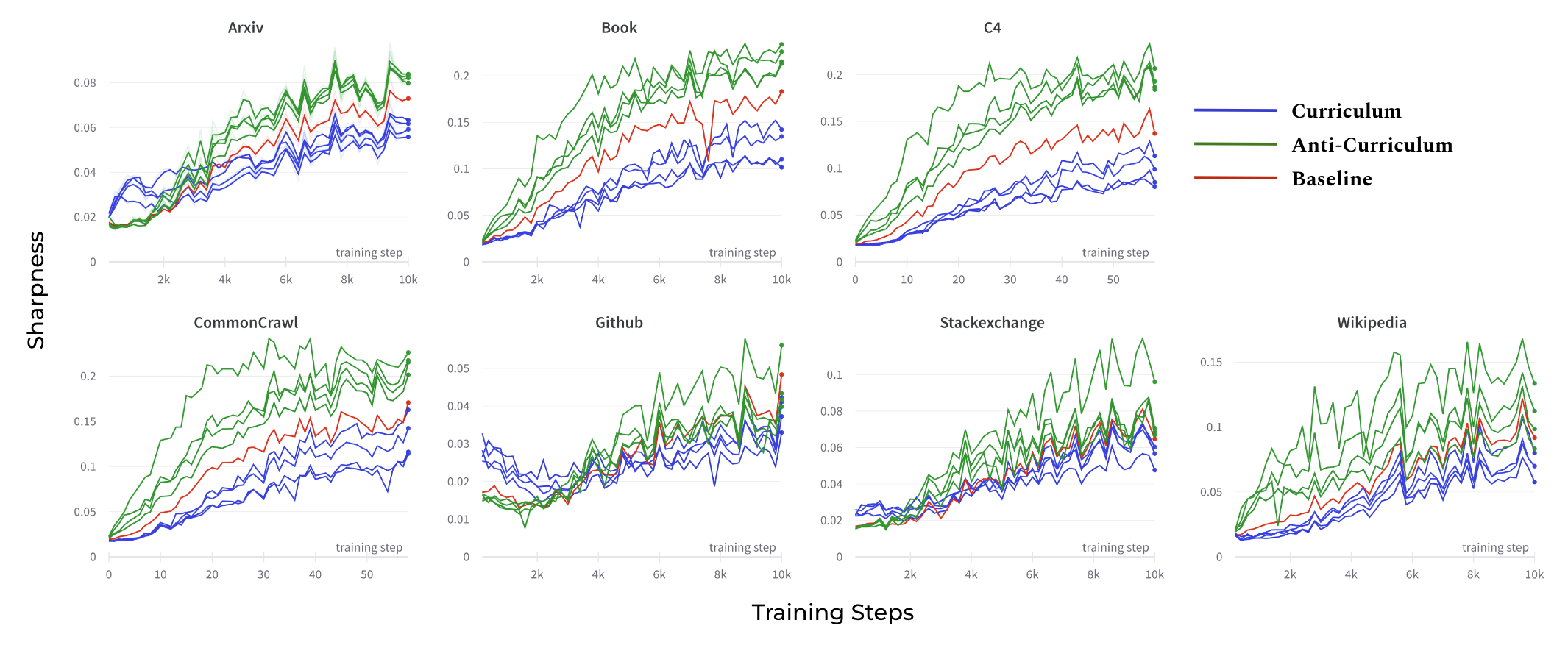}
        \caption{Network sharpness on RedPajama-1B. }
	\label{fig:domain-eigen-mix}
\end{figure}

\newpage